\documentclass[a4paper, 10 pt, conference]{cras} 
\usepackage[utf8]{inputenc}
\IEEEoverridecommandlockouts                       
\usepackage{mathtools}
\usepackage{geometry}
\usepackage{newtxmath,newtxtext}
\usepackage{authblk}
\usepackage{pgfplots}
\usepackage{todonotes}
\usepackage{subfig}
\usepackage{graphicx}
\usepackage{multirow}

\pgfplotsset{compat=newest} 
\pgfplotsset{plot coordinates/math parser=false} 
 
\title{\Large \bf
Forceps with direct torque control} 

\author{\large Zhuoqi Cheng}
\vspace{10pt}
\affil{\small\textit{Mærsk Mc-Kinney Møller Institute,
        University of Southern Denmark, 5230 Odense, Denmark}}

\begin{document}

\maketitle
\thispagestyle{empty}
\pagestyle{empty}

\section*{INTRODUCTION}

Minimally Invasive Surgery (MIS) is a modern surgical approach that utilizes advanced techniques and specialized instruments to perform procedures with minimal damage to surrounding tissues. One commonly used tool in MIS is the laparoscopic instrument, which is inserted through small incisions in the body for tissue manipulation or dissection. 

Conventional laparoscopic forceps use the handle opening angle to control the jaw opening angle. A common limitation of laparoscopic instruments is the ambiguous haptic feedback, which prevents the user from feeling the actual texture or resistance of the tissue being grasped. Surgeons can only guess the amount of applied force through visual cues and proprioception. 
In procedures involving delicate soft tissues, such as vascular, neurosurgical, and gynecological surgeries, surgeons should exercise particular caution to avoid using excessive grasping force, which may lead to tissue damage.

Recent advancements in surgical technology have explored the integration of tactile sensing into laparoscopic instruments \cite{kim2015force}. Such design potentially enhances the user's haptic perception during MIS procedures. However, force-sensing robotic forceps also come with certain disadvantages including high cost, structural complexity, unreliable feedback, and challenges in sterilization procedures.

This study presents a conceptual design of laparoscopic forceps whose grasping torque can be directly controlled by the user. By integrating an adjustable constant torque mechanism (ACTM), the handle opening angle is converted to the grasping torque irrespective of the jaw opening angle.
More details regarding the design methodology can be found in \cite{cheng2023design}.

\section*{MATERIALS AND METHODS}
\subsection{Adjustable constant torque mechanism}

\begin{figure}
    \centering
    \includegraphics[width=7.5cm]{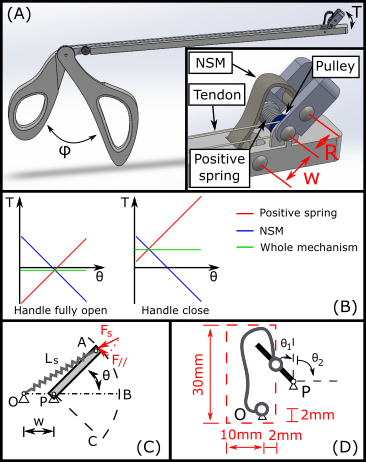}
    \caption{(A) Engineering design of the proposed forceps; (B) By paralleling a positive spring and a NSM, a constant torque output can be generated and controlled; (C) Mechanical sketch for NSM analysis; (D) The design boundary for obtaining an optimal compliant beam shape.}
    \label{fig:ACTM}
\end{figure}

ACTM is a passive compliant mechanism that outputs constant torque within the pre-defined displacement, while the torque magnitude can be controlled online.
It consists of a torsional spring of positive stiffness and a negative stiffness mechanism (NSM). The overall torque generated by the mechanism is the sum of both components. 

As shown in Fig. \ref{fig:ACTM}, we assume that the rotation centre of the crank is located at Point $P$ with radius $R$.
Point $B$ is used to denote the position during rotation and collinear with Point $O$ and $P$. $\theta$ is defined as the angle between $PB$ and the crank.
When the crank is rotated from one spring relax position $A$ to another spring relax position $C$, the torque on the crank $T$ generated by the deformation of the elastic element is a function of $\theta$.
\begin{equation}
    \label{eq:T}
    T(\theta) = F_{//}(\theta) \cdot R = \frac{F_S(\theta) \cdot w \cdot \sin \theta \cdot R}{L_S(\theta)}
\end{equation}
where $L_S(\theta)$ is the length of the elastic element.
By adjusting the pre-load of the positive spring as shown in Fig. \ref{fig:ACTM}(B), the output torque (green line) can be changed.

\subsection{Negative Stiffness Mechanisms}

According to Eq. \eqref{eq:T}, the stiffness of the elastic beam connecting $O$ and $A$ should be a non-linear function in order to make $T/\theta$ a constant. 
The design of NSM is an optimisation process involving Genetic Algorithm (GA) and Finite Element Method (FEM) to explore the effective variable space given the design constraints including geometrical, load, and displacement.
The optimisation procedure includes 
\begin{itemize}
    \item Generation: A pool of candidate beam shapes are generated. Each beam shape is defined via five key points $(x_i, y_i)$ within the predefined boundary. 
    \item Crossover: The crossover processing is operated on two random candidates with a possibility of 30\%. One of the three middle key points is randomly selected, and then they exchange their upper parts while the lower parts remain.
    \item Mutation: For the mutation operation, the middle key point multiplies a random number based on Gaussian distribution ($\mu=1, \sigma=0.01$).
    \item Selection: According to the fitness, each chromosome is assigned a different share of a roulette wheel. Then 40\% of the candidates are removed from the pool.
\end{itemize} 
The steps above are running in a loop until a candidate shape that can satisfy the stiffness function is found. 
It is worth mentioning that the fitness of the candidate in the $Selection$ step is based on the linear regression residual of its stiffness function, which is obtained from $FEM$.

After obtaining the best-fit beam shape, the section area of the compliant beam is optimised to amplify the output torque to match the required magnitude.

\subsection{Prototype}

A prototype of forceps was designed and fabricated as shown in Fig. \ref{fig:Exp}. 
The output torque for the forceps was set to be 0 to 30\,mNm, and the jaw length was set to be 20\,mm. The jaw opening angle was decided to be up to 90$^\circ$. 

A positive spring is required whose stiffness constant is 0.3\,mNm/$^o$. One end of the positive spring is fixed to the mobile jaw and the other end is to the pulley (Fig. \ref{fig:ACTM}). Tendon twines the pulley for controlling the pre-load on the positive spring. 

Also, the NSM is designed connecting the mobile jaw and the base jaw through pins. The offset distance $w$ was set to be 12\,mm, and $R$ to be 8\,mm for the purpose of making the forceps compact.
when the handle is fully open, the pre-load on the positive spring outputs a torque in the opposite direct with a magnitude slightly bigger than the output of the NSM. Thus, the combined torque is negative and drives the forceps jaw to open. When the handle opening angle is controlled smaller, the pre-load on the positive spring increases. The jaw closes when the output torque turns positive. Subsequently, the grasping torque increases with the handle open angle.

For the compliant beam design, the boundary condition was set to be 12\,mm$\times$30 mm and its relative position is shown in Fig. \ref{fig:ACTM}(D). An extra angle of $\theta_1$ = 45$^o$ was set as the pre-load period for the NSM before the negative stiffness appears. The following negative stiffness region $\theta_2$ = 90$^\circ$ was used as the operational displacement. The conditions were input into the optimisation algorithm and an appropriate design was generated. In addition, the section area of the compliant beam was increased to 2\,mm$\times$6\,mm in order to match the required stiffness value.
PLA material (Young’s modulus 3.45\,GPa and Poisson’s Ratio 0.39) was chosen for making the compliant beam, and Polyjet 3D printing technology was used to make the other mechanical parts. The simulation results of the elastic element indicate that the maximum von Mises stress at position $B$ is 69\,MPa which is smaller than the material's yield limit (106\,MPa).

\section*{EXPERIMENTS AND RESULTS}

\begin{figure}
    \centering
    \includegraphics[width=8cm]{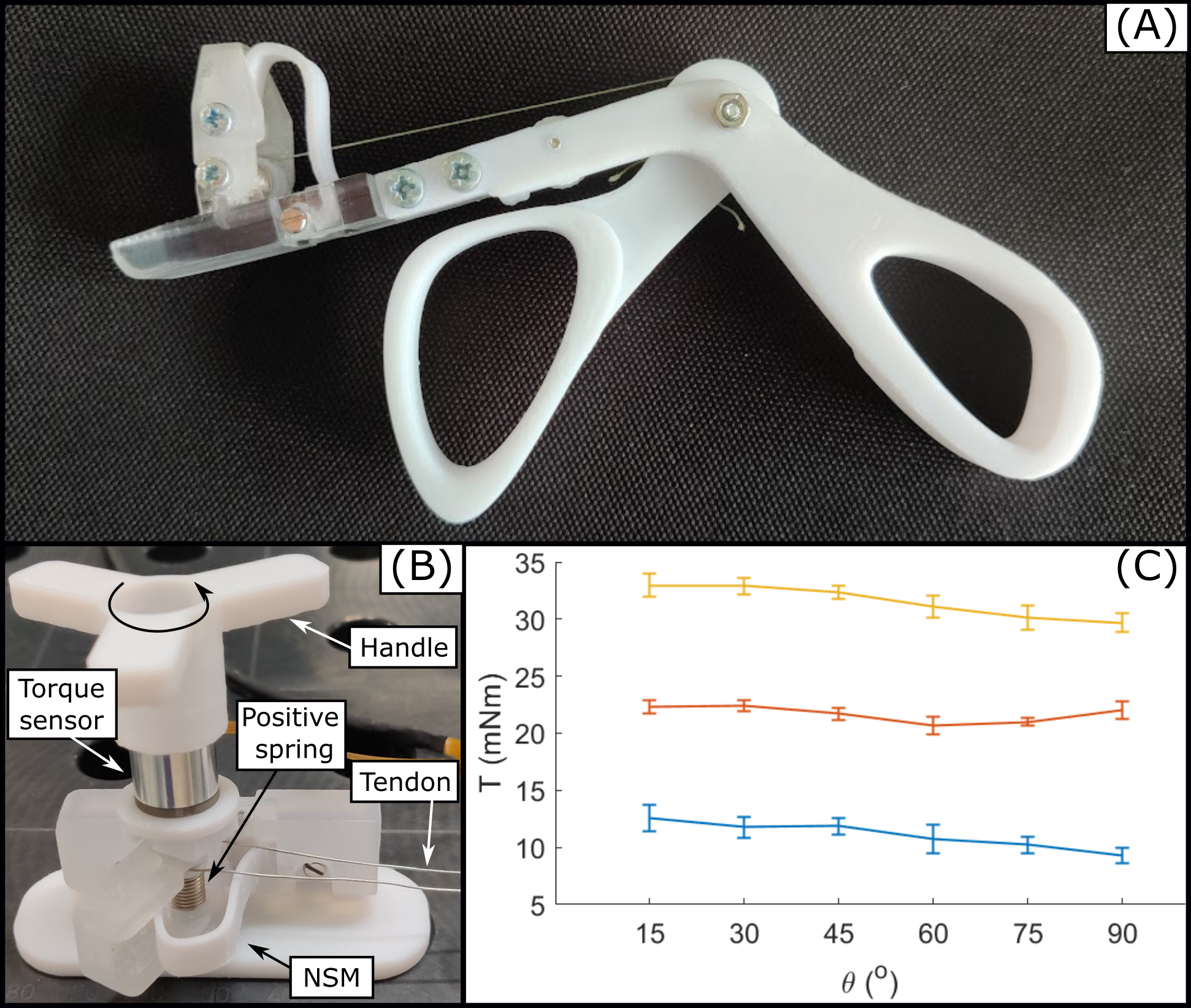}
    \caption{(A) Prototype of the proposed forceps; (B) Experimental setup for characterizing the output torque; (C) The output torques of the forceps with expected value of 10\,mNm, 20\,mNm, and 30\,mNm.}
    \label{fig:Exp}
\end{figure}

Fig. \ref{fig:Exp} shows the experimental setup. During the evaluation of the ACTM, a tendon was stretch to different distances for controlling the output torque. A 6 axial F/T sensor was connected between the mobile jaw and the handle. This allowed the torque on the mobile jaw to be measured.
During the experiments, the tendon was first stretched with a specific distance given the expected torque. Then the experimenter rotated the handle to open the mobile jaw from 15 to 90$^\circ$ in a step size of 15$^\circ$ manually. Then 1000 readings from the torque sensor were collected and sent to the connected laptop for processing.

Fig. \ref{fig:Exp}(C) shows the mean and standard deviation for all the results collected in each torque setting. The mean and standard deviation are found to be 11.08$\pm$1.44\,mNm, 21.67$\pm$0.89\,mNm, and 31.52$\pm$1.57\,mNm for the expected value of 10\,mNm, 20\,mNm, and 30\,mNm.

\section*{CONCLUSIONS AND DISCUSSION}

This study presents a new concept of laparoscopic forceps design and characterization.
Although there is offset between the results and the expected values, the constancy of the output torques of the designed forceps is demonstrated. Also, the results prove that the torque on the forceps can be changed online by controlling the handle opening angle.
The future research will focus on developing user interfaces that either restore haptic feedback or ensure secure tissue grasping.

\nocite{*}
\bibliographystyle{IEEEtran}
\bibliography{CRAS}

\end{document}